\documentclass[letterpaper, 10 pt, conference]{ieeeconf}  %

\usepackage{amsmath}
\usepackage{tikz}
\usetikzlibrary{arrows.meta}

\IEEEoverridecommandlockouts                              %

\usepackage{graphics} %
\usepackage{epsfig} %
\usepackage{mathptmx} %
\usepackage{times} %
\usepackage{amsmath} %
\usepackage{amssymb}  %
\usepackage{booktabs}
\usepackage{todonotes}
\usepackage[normalem]{ulem}
\usepackage{caption}

\title{\LARGE Self-Improving Autonomous Underwater Manipulation}

\usepackage{url}
\usepackage{hyperref}

\author{Ruoshi Liu$^{1}$, Huy Ha$^{1,2}$, Mengxue Hou$^{3}$, Shuran Song$^{1,2}$, Carl Vondrick$^{1}$ \\ %
$^{1}$Columbia University 
$^{2}$Stanford University
$^{3}$University of Notre Dame
  \\ \href{https://aquabot.cs.columbia.edu}{\url{aquabot.cs.columbia.edu}}
}

\begin{document}

\maketitle
\thispagestyle{empty}
\pagestyle{empty}

\begin{abstract}
Underwater robotic manipulation faces significant challenges due to complex fluid dynamics and unstructured environments, causing most manipulation systems to rely heavily on human teleoperation. In this paper, we introduce AquaBot, a fully autonomous manipulation system that combines behavior cloning from human demonstrations with self-learning optimization to improve beyond human teleoperation performance.
With extensive real-world experiments, we demonstrate AquaBot's versatility across diverse manipulation tasks, including object grasping, trash sorting, and rescue retrieval. Our real-world experiments show that AquaBot's self-optimized policy outperforms a human operator by 41\% in speed. AquaBot represents a promising step towards autonomous and self-improving underwater manipulation systems.
We will open-source both hardware and software implementation details.
\end{abstract}

\begin{figure}[htbp!]
    \centering
    \includegraphics[width=1.0\columnwidth]{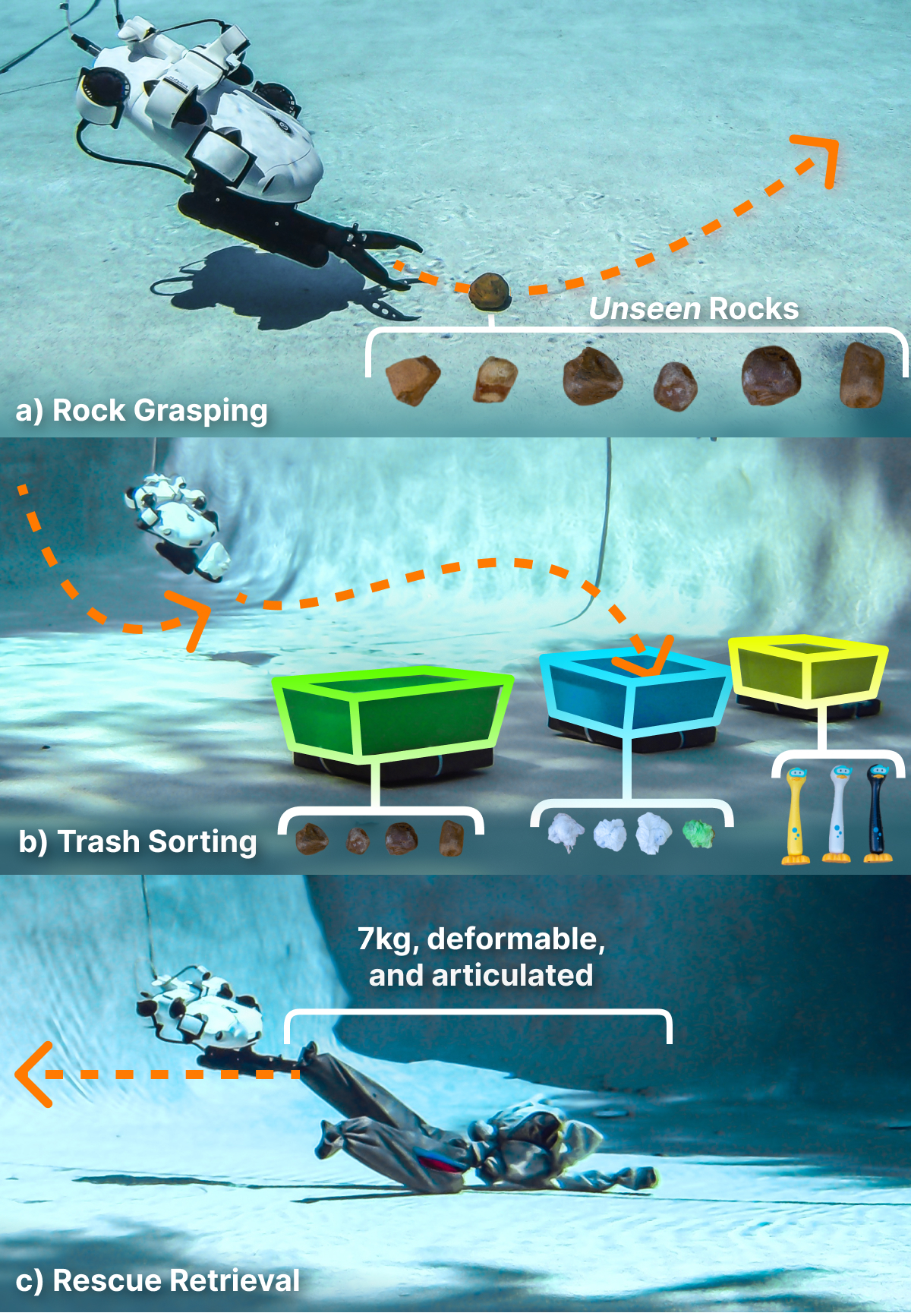}
    \caption{
        AquaBot combines behavior cloning with self-learning to optimize fully autonomous end-to-end visuomotor policies to achieve efficient manipulation skills across a wide range of tasks, including generalization to unseen objects (Rock Grasping), long horizon tasks (Trash Sorting) and robustness against large perturbations from unmodelled deformable and articulated objects (Rescue Retrieval).
    }
    \label{fig:teaser}
\end{figure}

\section{INTRODUCTION}

Despite high demand and continuous efforts, a fully autonomous manipulation system is still out of reach for most critical underwater applications. What makes underwater manipulation so hard?  In addition to the typical difficulties robotic manipulation systems face, underwater robots encounter unique challenges posed by high-dimensional and nonlinear fluid dynamics. Such complexity in modeling is further exacerbated by the wide variety of objects, tasks, and environments encountered underwater. This leads to significant challenges in controller design for underwater manipulation. As a result, many hand-engineered robotics manipulation systems fall short when attempting to explicitly model the complex dynamics of both the robot and its environment.

On the other hand, teleoperated systems \cite{7742315, stuart2017ocean, brantner2021controlling} benefit from human adaptability, allowing them to handle unpredictable conditions and execute complex tasks. However, teleoperation is not the perfect solution. First, its scalability is constrained by its dependence on human input. Furthermore, the system's performance is inherently capped by the operator's skill level, which is often suboptimal. For instance, operators usually operate the robot at low speeds when executing precise manipulation tasks, which trades off the system's performance for ease of teleoperation.

This raises a critical question: Can we distill the strategies demonstrated by human operators into a robust and reactive visuomotor policy that enables autonomous execution of manipulation tasks under complex dynamics? Furthermore, can the robot system continuously improve its learned policy through self-guided exploration and optimization, thereby exceeding the performance of human operators?

To answer this question, we propose AquaBot, a system designed to learn end-to-end visuomotor policies for fully autonomous underwater manipulation that can improve beyond human teleoperation performance through self-learning. The training process consists of two stages: In the first stage, we \textbf{distill} human adaptability into a closed-loop visuomotor policy. To do so, we record human demonstrations that teleoperate the robot to perform various manipulation tasks and then use the data to train a visuomotor policy. By shortening the policy's action horizon, we effectively increase the reactiveness of the policy, which is critical in dealing with unexpected underwater dynamics. In the second stage, we allow the robot to \textbf{accelerate} its learned behavior through self-guided optimization. In this step, we repeatedly execute the learned policy and use the execution time as a reward to accelerate the policy with a surrogate-based optimization algorithm. This step allows the system to further optimize those suboptimal parameters (e.g., execution speed) in the human demonstration data.

Through extensive real-world experiments, we empirically found that our learning-based system offers several major advantages when compared with classical underwater robot controllers:

\begin{itemize}
    \item \textbf{Versatility.} We apply the same method across three different challenging manipulation tasks, including object grasping, trash sorting, and rescue retrieval.
    \item \textbf{Simplicity.} Unlike prior systems, our method handles perception, dynamics modeling, motion planning, and control within a single end-to-end visuomotor policy.
    \item \textbf{Self-improving.} Our base policy learned from demonstrations can continue to improve as the robot accumulates more experience in the field.
\end{itemize}

Our experiments have shown that end-to-end visuomotor policies can handle multiple manipulation tasks with high accuracy. With only 120 trials, our accelerated policy outperforms the human operator by 41\% and base policy by 68\% in manipulation speed.

\section{RELATED WORK}
\subsection{Autonomous Underwater Manipulation}
Prior works in underwater manipulation have looked at various components of a system, including manipulator and end-effector design, dynamic modeling, model-based controllers, motion planning, and perception systems~\cite{huang2020review, aldhaheri2022underwater}.

\textbf{Dynamics.}
Mathematical models are extensively used in the control of robotic systems. Applying these controllers for on-land robots directly to underwater robots poses significant problems due to the hydrodynamic effects. The hydrodynamic effects of simple two-link manipulators were studied in~\cite{724376, 5978927, gumucsel2011modelling}. Further studies on modeling dynamics of vehicle-manipulator systems were done in~\cite{7002989, 5978927} to study the dynamic coupling effectors between the vehicle and attached manipulators. Another innovative line of work focuses on modeling the dynamics of bio-inspired underwater swimming manipulators (USM)~\cite{6907522, SVERDRUPTHYGESON201681, 8121980, fairchild2023physics}.

\textbf{Control.}
Prior works have also extensively studied the design of controllers to cope with the disturbances and uncertainties introduced by water environments. A line of work focuses on developing control laws for underwater robotic control%
~\cite{BARBALATA2018150, BARBALATA201544,  HAUGALOKKEN20181}. More recently, deep reinforcement learning (DRL) was used in robotic control~\cite{mnih2015human, tassa2018deepmind} and inspired many works to use DRL for underwater robotic control~\cite{jadhav2017rov, Carlucho_2020, yang2021prediction}. Another line of work focuses on learning a Model Predictive Control (MPC) controller with neural networks~\cite{8729801, salloom2020adaptive, carlucho2021adaptive}.

\textbf{Perception.}
Prior works have explored topics including perception~\cite{oubre2021datadrivencontrollersneedperception}, sensing~\cite{chee2023learnest}, and multi-sensor fusion~\cite{rahman2022svin2}. More recently, with the rapid development of computer vision, such as techniques like SLAM~\cite{davison2007monoslam, matsuki2024gaussian}, neural rendering~\cite{kerbl20233d, mildenhall2021nerf}, 3D reconstruction~\cite{liu2023zero, wang2024dust3r}, many research works have started to explore data-driven underwater perception systems~\cite{tang2024uwnerf, zhang2023beyond}. 

\begin{figure}[t]
    \centering
    \includegraphics[width=1.0\columnwidth]{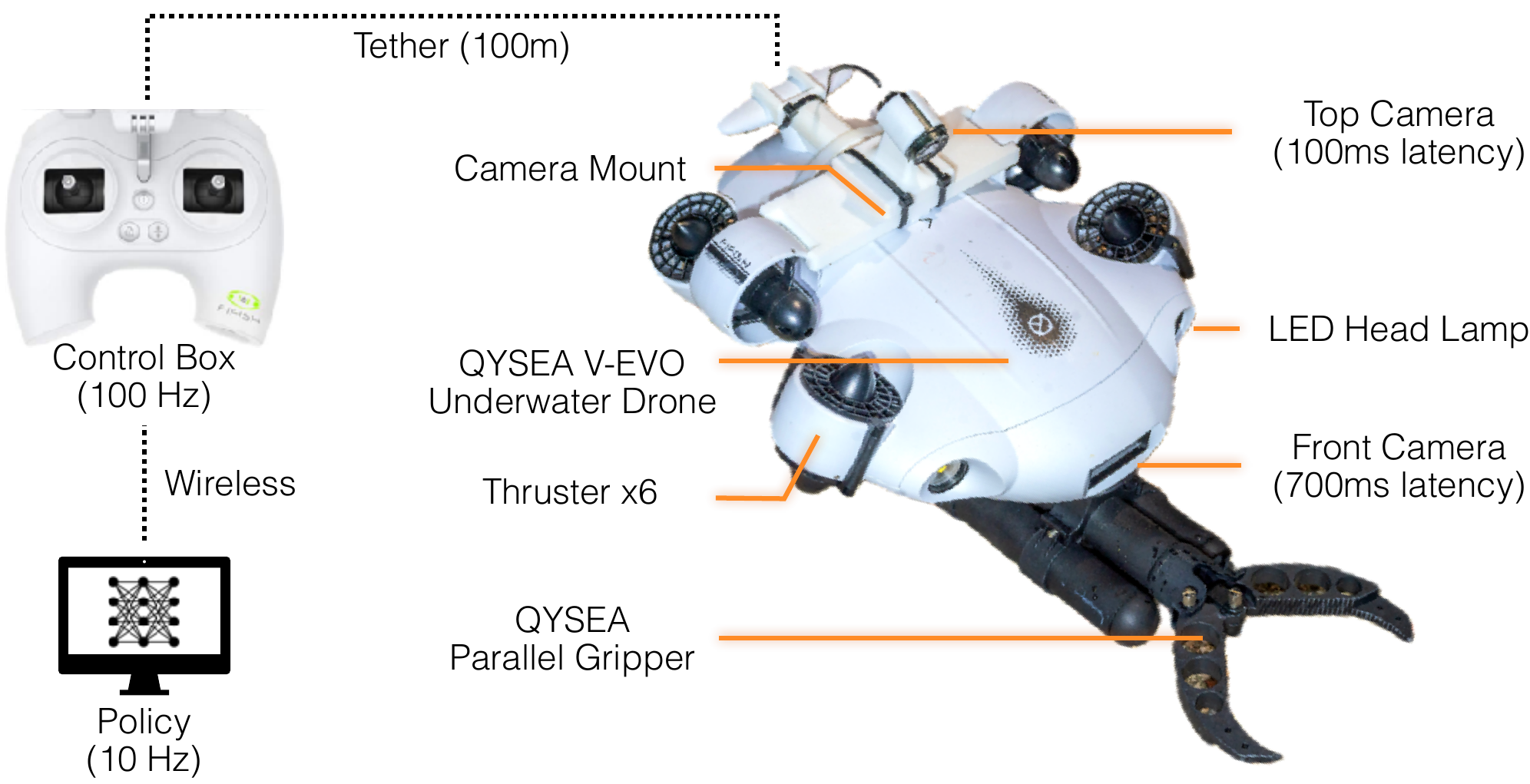}
    \caption{Our accessible hardware platform (\$2000 USD) consists of 2 cameras and a parallel jaw gripper for research and development of underwater visuomotor policy learning.}
    \label{fig:hardware}
\end{figure}

\begin{figure*}[t]
    \centering
    \includegraphics[width=1.0\textwidth]{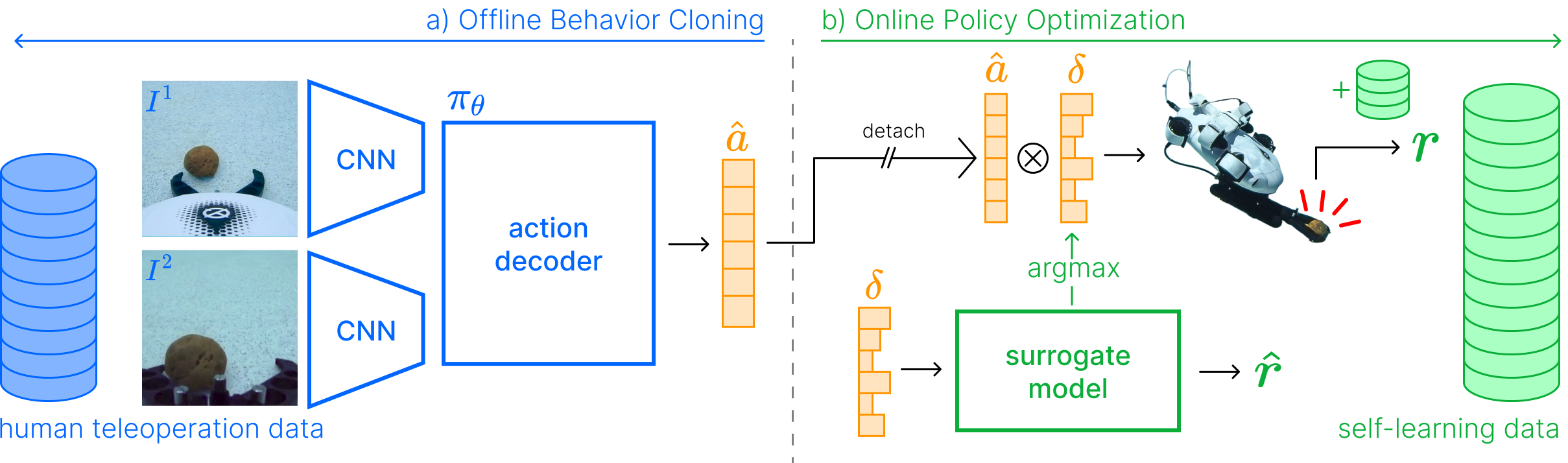}
    \caption{\textbf{Learning Framework.} 
    In the first stage \textbf{(a)}, we train our base policy by learning from human demonstrations from offline data.
    In the second stage \textbf{(b)}, we roll out the behavior-cloned policy to collect more self-learning data to learn a surrogate model in an online fashion, which optimizes the motor speed $\delta$ in an online fashion.}
    \label{fig:method}
\end{figure*}

\subsection{Sensorimotor Policy}
Sensorimotor policy is a decision-making model that chooses actions to control a dynamical system based on sensory data. Such actions could be, for example, the motor torque commands of a robot, and the sensory data could come from on-robot or external sensors such as cameras, sonar, thermostats, and water pressure sensors.
Reinforcement learning (RL) has been one of the main avenues of research towards learning visuomotor policy for manipulation~\cite{6301026, Kober2014, 6095096}. More recently, deep reinforcement learning (DRL) uses deep neural networks to parameterize a policy that can be optimized under a reinforcement learning algorithm~\cite{levine2016end, schulman2017proximal, haarnoja2018soft, akkaya2019solving}. Another line of works focuses on learning a world model as a differentiable world simulator, in which manipulation could be performed through optimization-based planning or a policy can be learned through reinforcement learning~\cite{ha2018world, hafner2019learning, hafner2019dream, wu2023daydreamer, liang2024dreamitate}. Dynamic action primitives were extensively studied to enhance a policy's ability to perform manipulation in a dynamic environment~\cite{zeng2020tossingbot, xu2022dextairity, ha2022flingbot, chi2024iterative, liu2024paperbot}.

\section{APPROACH}

\subsection{Hardware / Software System}
Our hardware platform (Fig.~\ref{fig:hardware}) is built around the QYSEA FIFISH V-EVO underwater drone which is a cheap (\$1600) and accessible ROV drone with six thrusters providing 6 DoF torque and force control. The drone also comes with a parallel gripper that can be attached externally to the drone body. The front camera has a relatively high latency (700 ms). Therefore, we externally mounted a low-cost (\$170) waterproof streaming camera with 100 ms latency on top of the robot body, also adding a stereo view for manipulation. A 3D-printed camera mount was fixed to the robot body. The movement of the robot and its gripper is controlled by a tethered control box with a 100 Hz control frequency. For autonomous manipulation, we use a software SDK to send control signals to the robot through the control box and receive proprioceptive data through the same connection. Additionally, we mounted two external cameras at two corners of the pool for real-time localization of the robot. The detected 3D position, plus the internal IMU sensor and compass, provide a full 6 DoF robot pose in the global coordinate system, which we use for navigation and reset. Note that this is a proxy for existing localization and navigation methods, a well-studied area in marine robotics with extensive prior works.

\subsection{Manipulation Policy with Behavior Cloning}
Since we cannot precisely detect the position and orientation of the robot underwater, we perform force and torque control instead of position control (commonly used for on-land manipulation~\cite{chi2023diffusion, zhao2023learningfinegrainedbimanualmanipulation, lee2024behavior}), as shown in Fig.~\ref{fig:method}(a). We implemented a teleoperation system with an Xbox controller to collect human demonstrations for different tasks. During demonstration collection, we record visual data at 10 Hz and control data at 30 Hz. Each recorded action is an 8D vector composing the 3 Cartesian directions, 3 rotational directions, and open/close gripper movement.

With demonstration data collected, we follow the typical behavior cloning methods~\cite{levine2016end, chi2023diffusion, zhao2023learningfinegrainedbimanualmanipulation} for learning an end-to-end visuomotor policy. For the visual encoder, we use a convolutional neural network (CNN) to obtain a feature vector for each image. We use two separate visual encoders for each camera (top camera + front camera), each with an observation horizon of 2. The features were concatenated as the input to a multi-layer perceptron (MLP) to predict the 8D action vector, which is supervised with an MSE loss (empirically better than L1). Formally, we optimize for

\begin{equation}
    \underset{\theta}{\arg\min}\, \mathcal{L} = \mathop{\mathbb{E}}_{a, I} \left\| a - \pi_{\theta}(f(I^{1}), g(I^{2})) \right\|_2
\end{equation}

where $\pi_\theta$ is the learned policy with parameters $\theta$, $a$ is ground truth action vector, $f, g$ are vision encoders, and $I^{1}, I^{2}$ are observation from the two camera stream.
When other action decoders, such as diffusion and transformer-based CVAE, are used, we apply their corresponding loss function for supervision. During deployment, our policy inference time is well below 100ms, so we are able to perform 10 Hz real-time control of the robot using the learned policy.

\begin{figure*}[t]
    \centering
    \includegraphics[width=1.0\textwidth]{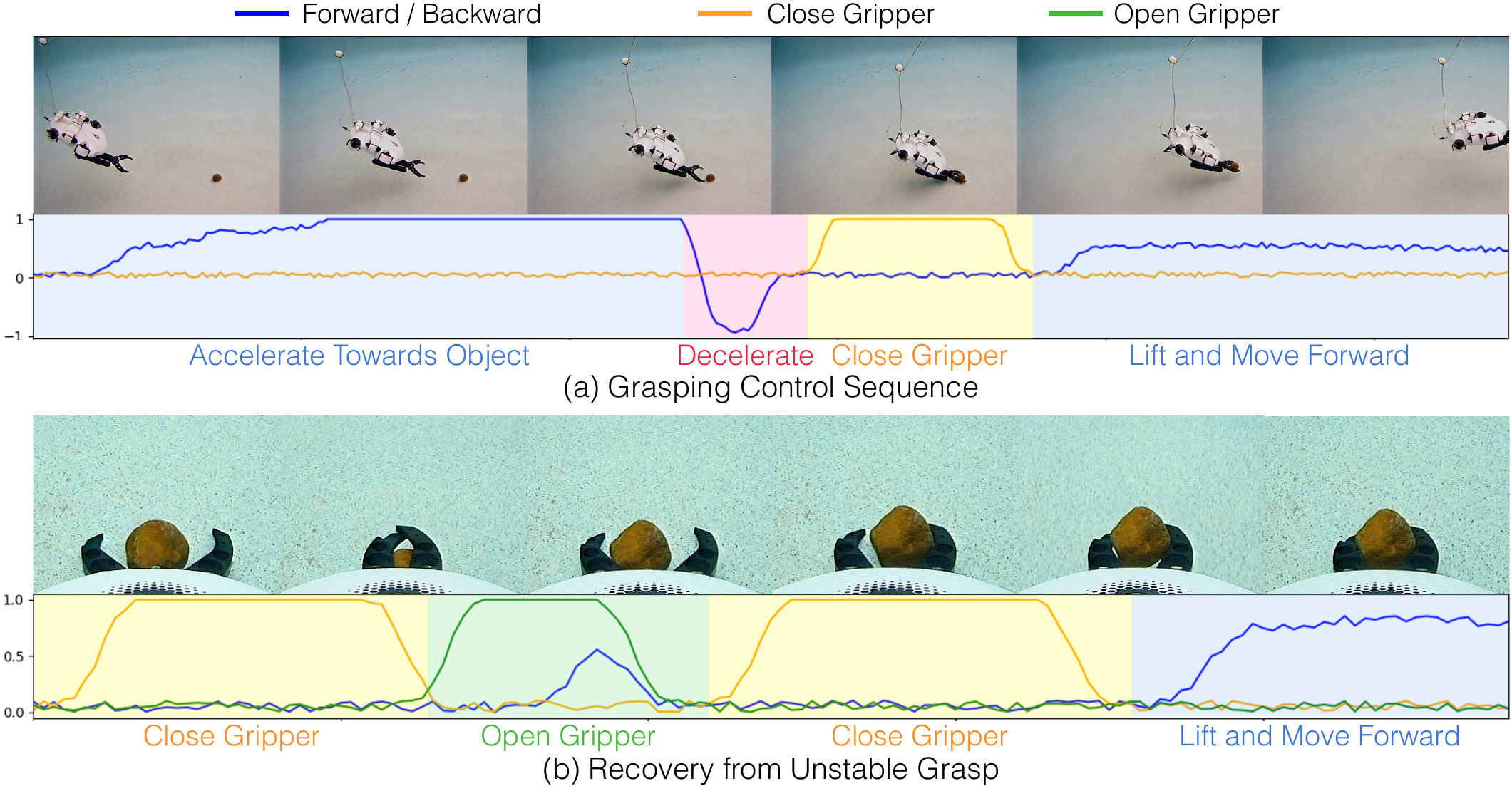}
    \caption{\textbf{Dynamic and Robust Manipulation.} 
    We plot policy outputs below their corresponding third-person views.
    \textbf{(a)} shows how the robot decelerates by applying backward forces when it is close to the object, demonstrating proficiency in underwater dynamics. 
    \textbf{(b)} shows how the policy will retry after unstable grasps, demonstrating robustness.
    }
    \label{fig:qualitative}
\end{figure*}

\textbf{Implementation Details.}
Both of the camera inputs were resized to 224x224. We use a ResNet-18 with the last residual block removed and replaced with a spatial-softmax layer~\cite{levine2016end} to obtain a 512x2 feature vector for each input image. We concatenate all feature vectors from an observation horizon of =2 for both cameras as the input to the 3-layer MLP with LeakyReLU activation layers and a hidden dimension of 64. For each task, we train on the demonstrations for 50 epochs with a batch size of 32 and an lr of 1e-4. For DP and ACT, we use hyperparameters in the original official implementation. During deployment, policy inference and real-time localization are run on two RTX A6000 GPUs. A 32-core AMD Threadripper CPU was used to handle the real-time decoding of camera streams and other operations.

\subsection{Self Learning for Policy Acceleration}
Due to the limitation of teleoperation systems and the human's lack of underwater motor skills, human demonstrations are likely to be sub-optimal. This could manifest in several ways, such as the complexity of demonstrations, manipulation efficiency, and the inability to predict effects caused by water currents. A visuomotor policy learned from behavior cloning can be significantly improved through self-learning, where a robot performs a task by itself and learns to adjust its behavior to optimize for a predefined reward. Since most policies are trained once and deployed numerous times. A self-learning method can benefit from the continual accumulation of robot deployment data.

\textbf{Safe Environment for Self Learning.} Self-learning is typically costly to set up on land due to safety concerns. Luckily, water provides a safety buffer for the robot. In addition, the one-piece design of the underwater drone is physically robust. These conditions create a safe environment for our robot to perform self-learning without engineering many safety constraints. Since our policy outputs a continuous 6 DoF force/torque control signal, we can learn to accelerate the policy by learning a scaling parameter for each control dimension, where the objective is to complete a manipulation task in the shortest time possible. 

\textbf{Learning to Accelerate.} To solve this non-convex gradient-free optimization problem, we adopted the surrogate-based optimization algorithm proposed by~\cite{liu2024paperbot} as illustrated in Fig.~\ref{fig:method}(b). 
Consider a reward we are learning to optimize $r$, a BC-trained policy $\pi$, and our goal is to maximize the reward by learning the optimal scaling parameters $\delta$, which is time-invariant.
To optimize a neural surrogate model that maps from $\delta$ to $r$, we use an epsilon-greedy exploration strategy as follows. 
During an exploration episode, we randomly sample a $\delta$ from a uniform distribution. 
During an exploitation episode, we use the neural surrogate model to optimize for the best parameter $\delta$.
At the end of the episode, we measure $r$, add $\{\delta,r\}$ to the self-learning dataset, and use all previously self-collected data to train the surrogate model. 
The implementation for automatic reward evaluation and reset for each task can be found in~\ref{exp:self-learn}.

\textbf{Implementation Details.} The surrogate model used is a 3-layer MLP with an input dimension matching the corresponding parameter space, hidden dimensions of 512, and a scalar output for reward. Given a dataset of pairs of input speed parameters and execution time as a reward, we train the network for $1000$ iterations with a batch size of 8. We use an AdamW optimizer with a learning rate of 0.01 and weight decay of 0.1 to optimize for the Huber loss (smoothed L1 loss) between the predicted reward and ground truth. Because the surrogate model training and inverse optimization combined take less than 2 seconds on an NVIDIA RTX A6000 GPU, which is far from being the bottleneck during experiments, we perform these two steps before each trial.

\section{EVALUATION}

\subsection{Behavior Cloning}

\begin{table}[h!]
\centering
\caption{Object Grasping Performance}
\begin{tabular}{lccc}
\specialrule{1.2pt}{1pt}{1.2pt}
\textbf{Metric} & \textbf{MLP} & \textbf{DP} & \textbf{ACT} \\
\hline
\textbf{Success Rate} & 20/20 & 8/20 & 0/20 \\
\textbf{Average Time (s)} & 13.52 & 45.53 & N/A \\
\textbf{Standard Deviation (s)} & 3.59 & 18.53 & N/A \\
\hline
\end{tabular}
\label{tab:policy_comparison}
\end{table}

\subsubsection{\textbf{Task 1: Object Grasping}}

Grasping is a classic robotic task extensively studied in prior works.
However, water makes the task significantly harder. Many on-land grasping policies rely on precise positional control and long-horizon motion planning, which are both difficult to achieve. In addition, with disturbances and uncertainty created by water, a grasping policy needs to constantly counter these forces to navigate towards the target grasping pose and stabilize the robot pose while closing the gripper. Despite its difficulty, a human operator, with several hours of training, is able to control the robot with an Xbox controller to perform reliable grasping. Therefore, we aim to distill such manipulation skills demonstrated by humans into a policy with behavior cloning. An illustration of these tasks can be found in Fig.~\ref{fig:teaser}(a). For this task, we collected 492 human demonstrations.

\textbf{Findings:} for this task, we evaluate the performance of learned policy with three different policy decoder architectures, including vanilla MLP, Diffusion Policy (DP)~\cite{chi2023diffusion}, and Action Chunk Transformers (ACT)~\cite{zhao2023learningfinegrainedbimanualmanipulation}. \uline{We found that MLP policy yields the highest performance and manipulation efficiency, outperforming both DP and ACT.} Further analyzing the policy behavior, we found that the biggest failure mode of both DP and ACT is the gripper motion. With a short action horizon, DP fails to generate continuous gripper closing and opening motions, causing the grasping to be unstable or unsuccessful. After training, ACT fails to generate any gripper motion.

\begin{table}[h]
\centering
\caption{Trash Sorting Performance}
\begin{tabular}{lcccccc}
\toprule
\textbf{Object Category} & \multicolumn{2}{c}{\textbf{Toy}} & \multicolumn{2}{c}{\textbf{Rock}} & \multicolumn{2}{c}{\textbf{Plastic}} \\
\cmidrule(lr){2-3} \cmidrule(lr){4-5} \cmidrule(lr){6-7}
\textbf{Policy Architecture} & \textbf{MLP} & \textbf{DP} & \textbf{MLP} & \textbf{DP} & \textbf{MLP} & \textbf{DP} \\
\midrule
\textbf{Success Rate} & 10/10 & 8/10 & 10/10 & 4/10 & 9/10 & 9/10 \\
\bottomrule
\end{tabular}
\label{tab:trash_sorting}
\end{table}

\begin{figure}[t]
    \centering
    \includegraphics[width=1.0\columnwidth]{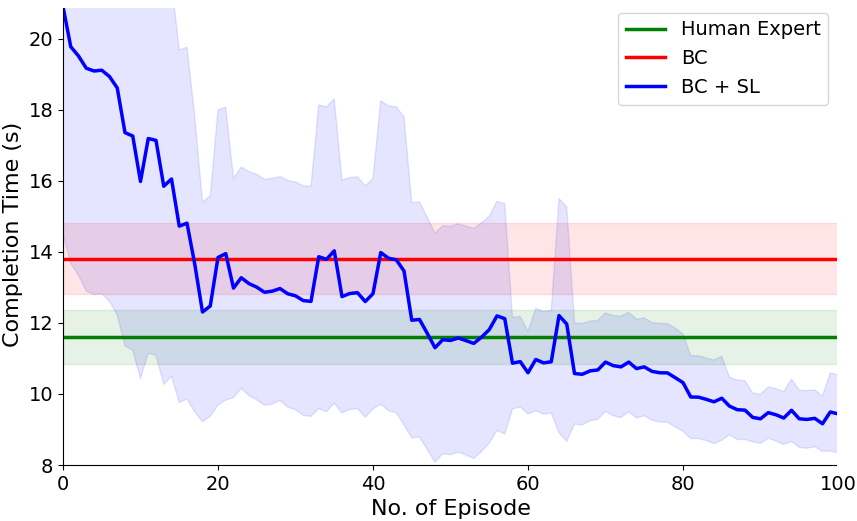}
    \caption{By self-learning (SL), AquaBot learns to accelerate a manipulation policy learned from Behavior Cloning (BC) through trial-and-error. After only 100 iterations, it can perform the same manipulation task more efficiently than vanilla BC policy and human experts.}
    \vspace{-0.3cm}
    \label{fig:self-learning}
\end{figure}

\subsubsection{\textbf{Task 2: Trash Sorting}}

Motivated by the potential application of autonomous ocean trash collection with robots, we designed a trash sorting task where the robot learns to pick up an object, predict its object classes, and place it in the corresponding bin. This task demonstrates the system's ability to perform long-horizon tasks. We collected a total of 527 demonstrations from human teleoperation to learn a grasping policy for 3 object categories as shown in Fig.~\ref{fig:teaser}, each with six different object instances. In addition, we train a classifier model to predict the object category after grasping it successfully. Finally, we use two external cameras to perform real-time localization and use a PID controller to navigate the robot and place the object in the corresponding bin. An illustration of this task can be found in Fig.~\ref{fig:teaser}(b).

\textbf{Findings:}
As shown in Tab.~\ref{tab:trash_sorting}, we found that both MLP and DP perform well on grasping toys and plastic bags, but MLP performs better than DP for grasping rocks. \uline{Through this task, we found that a single system can autonomously perform grasping, sorting, and placing of objects with a variety of appearances, material properties, geometry, and mass.}

\begin{table}[h]
\centering
\caption{Success Rate Comparison for Rescue Retrieval}
\begin{tabular}{lcc}
\toprule
\textbf{Policy} & \textbf{MLP} & \textbf{DP} \\
\midrule
\textbf{Success Rate} & 5/10 & 3/10 \\
\bottomrule
\end{tabular}
\label{tab:people_rescuing}
\end{table}

\subsubsection{\textbf{Task 3: Rescue Retrieval}}
Third, we teach our robot to perform a rescue retrieval task where a robot needs to grasp an object larger and heavier than itself and drag it to a target area, as shown in Fig.~\ref{fig:teaser}(c). In the experiment, the humanoid object weighs 6.8kg, which is significantly heavier than the robot's own weight of 3.8kg. For this task, we collected 100 demonstrations and found that MLP achieves slightly better performance than DP. \uline{Through this task, we found that underwater robots can manipulate objects much larger and heavier than their own body due to the presence of buoyancy.}

\begin{table}[h]
\centering
\caption{Performance vs. Action Horizon}
\begin{tabular}{lcccc}
\toprule
\textbf{Action Horizon} & \textbf{1} & \textbf{2} & \textbf{4} & \textbf{8} \\
\midrule
\textbf{Average Time (s)} &  &  &  &  \\
\quad MLP & 13.52 & 28.39 & 41.65 & 55.15 \\
\quad DP & 45.53 & 49.10 & 54.04 & 56.77 \\
\midrule
\textbf{Success Rate} &  &  &  &  \\
\quad DP & 20/20 & 16/20 & 10/20 & 4/20 \\
\quad Diffusion & 8/20 & 6/20 & 3/20 & 2/20 \\
\bottomrule
\end{tabular}
\label{tab:action_horizon}
\end{table}

\textbf{Action Horizon.} Water introduces significant uncertainty in robot movement, resulting in low repeatability and making pose stabilization challenging for underwater vehicles. Recent work on behavior cloning~\cite{chi2023diffusion, zhao2023learningfinegrainedbimanualmanipulation} has shown that action chunking—predicting and executing long action sequences—improves manipulation success and sample efficiency on land. However, in underwater environments, where currents increase movement uncertainty, shorter action horizons are more effective. \uline{Our ablation studies (Tab.~\ref{tab:action_horizon}) confirm that reducing the action horizon improves both success rate and policy efficiency.}

\textbf{Policy learns control and recovery behavior.} Fig.~\ref{fig:qualitative}(a) highlights our policy learns to control. During grasping, the robot first accelerates toward the object and then applies reverse forces to decelerate and prevent overshooting or colliding with the object, showing the policy's ability to perform both manipulation and robot control concurrently. Additionally, when failures occur, such as unstable grasps, the closed-loop policy quickly recovers, further illustrating the advantages of a shorter action horizon, as shown in Fig.~\ref{fig:qualitative}(b).

\subsection{Self Learning for Policy Acceleration} \label{exp:self-learn}
\begin{figure}[t]
    \centering
    \includegraphics[width=1\columnwidth]{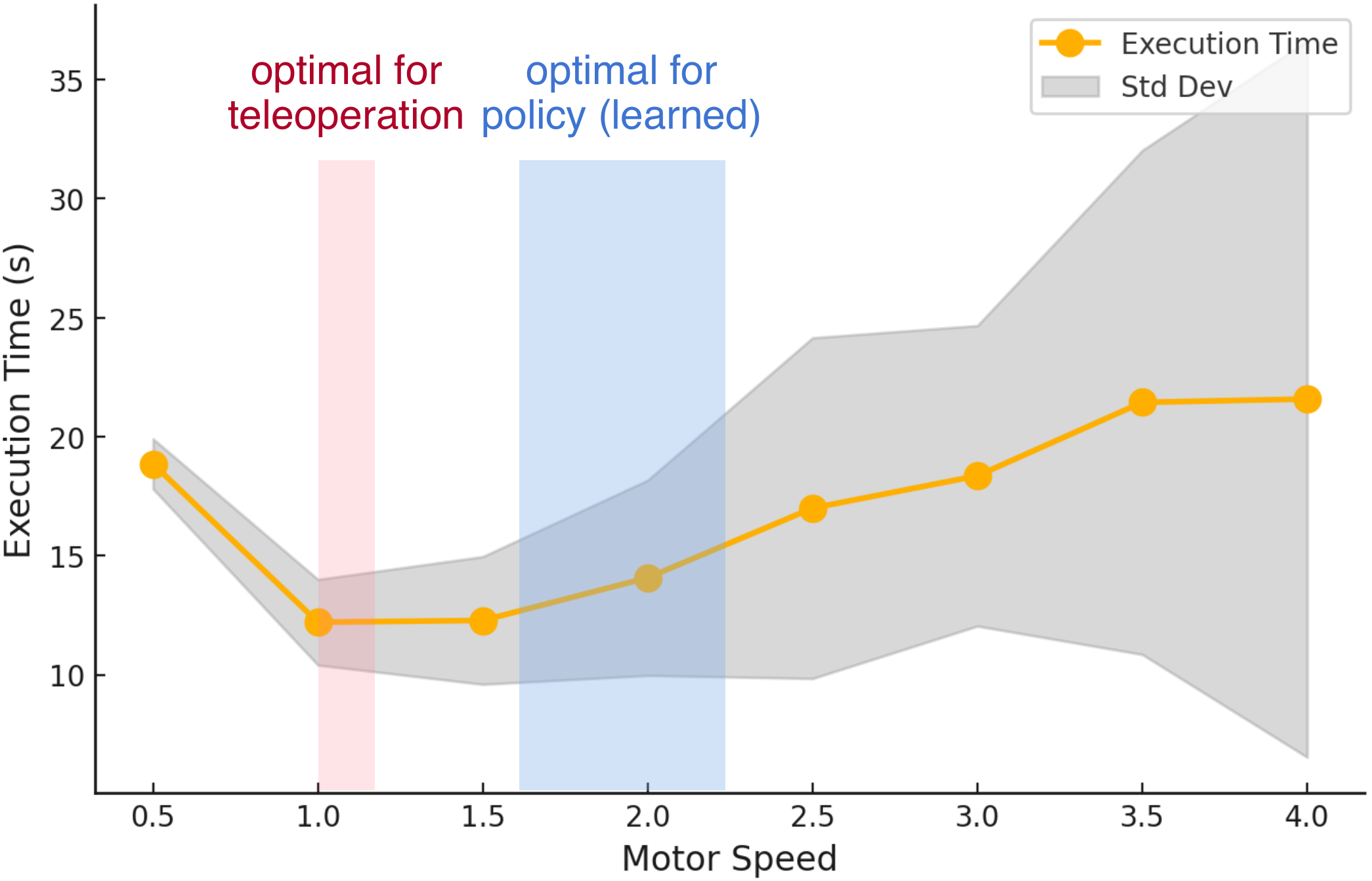}
        \caption{\textbf{Human Manipulation Efficiency vs. Motor Speed.} High motor speed leads to unstable manipulation and mistakes, and low motor speed leads to slow robot movements, both hurting the manipulation efficiency. There exists a sweet spot for the motor speed, balancing stability and speed.}
        \vspace{-0.1cm}
    \label{fig:human_efficiency}
\end{figure}

A key limitation of behavior cloning is that the policy learned from human demonstrations is inherently restricted by the performance of the human operator. During teleoperation, we observed that increasing motor speed leads to a higher rate of errors by the operator, as shown in Fig.~\ref{fig:human_efficiency}. 
Our self-learning algorithm is designed to search for the optimal combination of speed parameters through trial and error. To validate this approach, we conduct self-learning experiments on object-grasping tasks, demonstrating its effectiveness in improving task performance.

\textbf{Experimental Protocol.} With everything fully automated, we perform 120 episodes of policy rollouts to optimize a 5D speed vector, including (forward/backward, pan left/right, up/down, yaw, and pitch). We leave out the rolling action due to its irrelevance to the grasping task. We plot the results of self-learning in Fig.~\ref{fig:self-learning}, with the y-axis representing the task completion time of each episode and the x-axis representing the number of episodes conducted so far. We calculate the standard deviation for each episode from samples of moving window size of 20 and plot it as the error bars. For comparison, we also plotted the mean and standard deviation of the human teleoperation performance and BC policy with base speed, both calculated from 10 trials. The speed parameters are sampled uniformly random from $[0.5, 3]$.

\textbf{Assessment of Success.} To automate self-learning in the real world, we need to automatically predict whether a task is completed. For grasping, we check for 3 conditions:
\begin{itemize}
    \item the policy outputs a gripper closing signal or not
    \item the gripper is fully closed or not
    \item the gripper is moving or not
\end{itemize}
If the \textit{policy is controlling the gripper to close} and \textit{the gripper is not fully closed} and \textit{the gripper is not moving} are all true for 1 second, we decide that the grasping has been successful. One may argue that such criteria may lead to adversarial behavior where the robot finds non-target objects to grasp, which is often encountered during reinforcement learning~\cite{mnih2015human}. However, since we are not changing the parameters of the base policy, we never observe such adversarial behavior during our experiments.

\textbf{Reset Mechanism.} We use localization and navigation systems based on the external cameras to reset robot and object positions before each episode. During each episode, we start with the object already being grasped by the robot and navigate the robot towards a predefined starting position. Upon reaching the starting position, the robot releases the object on the ground. At the same time, the robot moves backward and upward for 2 seconds, after which the robot starts being controlled by the policy. We measure the time it takes for the robot to successfully grasp the object. The aforementioned criteria decide the success of grasping.

\begin{table}[t]
\centering
\caption{Learned Speed Generalizes to Other Tasks}
\begin{tabular}{lcc}
\specialrule{1.2pt}{1pt}{1.2pt}
  & \textbf{Object Sorting} & \textbf{Rescue} \\ & \textbf{(per object)} & \textbf{Retrieval}\\
\hline
\textbf{BC}       & 37.47s & 41.23s \\
\textbf{BC + SL} & 31.34s & 33.54s \\
\hline
\end{tabular}
\label{tab:sorting_retrieval_comparison}
\end{table}

\textbf{Result Analysis.} Fig.~\ref{fig:self-learning} shows that the manipulation efficiency from self-learning was initially significantly worse than that of the BC policy human baseline. This is expected because randomly sampled speed parameters from the range likely lead to worse behavior by either running too fast, which causes instability or moving too slowly to finish the task within the time limit. As more episodes are conducted, the surrogate model finds an optimal trade-off between efficiency and stability, thus able to obtain parameters that gradually improve the manipulation efficiency. \uline{Within only 120 trials (only 100 were shown due to a moving window size of 20), self-learning is able to find speed parameters leading to manipulation behavior that's significantly more efficient than the baselines, outperforming the human baseline by 41\% and BC baseline by 68\%}. After obtaining the optimal motor speed parameters, we apply them to other tasks and see a 19.6\% and 22.9\% improvement for trash sorting and rescue retrieval, respectively. This shows the improvement obtained from self-learning is not task-specific.

\section{CONCLUSIONS}
In this paper, we presented a self-learning framework for autonomous underwater manipulation. We first demonstrated that an end-to-end visuomotor policy can be learned from human demonstrations to control an underwater robot to perform a diverse set of manipulation tasks. We then apply self-learning to accelerate the learned policy by optimizing a speed parameter for each action dimension through trial and error. The final accelerated policy achieves much higher manipulation efficiency than both the BC policy and the human operator. We believe behavior cloning will be an excellent avenue for achieving robust autonomous underwater manipulation in the wild, and self-learning will continue to improve these manipulation skills to achieve superhuman performance. Finally, we hope to bridge the robot learning and marine robotics community, bring learning methods to underwater manipulation, and inspire more robot learning researchers to explore applications in underwater robotics.

\section{ACKNOWLEDGEMENTS}
We would like to thank Cheng Chi, Aurora Qian, Yunzhu Li, Zeyi Liu, Matei Ciocarlie, and Xia Zhou for their helpful feedback. We would also like to acknowledge the technical support from QYSEA. This work is supported in part by NSF Award \#2143601, \#2037101, and \#2132519, \#1925157, Sloan Fellowship. The views and conclusions contained herein are those of the authors and should not be interpreted as necessarily representing the official policies, either expressed or implied, of the sponsors. 

\bibliographystyle{IEEEtran}
\bibliography{IEEEexample, IEEEfull}

\begin{thebibliography}{10}
\providecommand{\url}[1]{#1}
\csname url@rmstyle\endcsname
\providecommand{\newblock}{\relax}
\providecommand{\bibinfo}[2]{#2}
\providecommand\BIBentrySTDinterwordspacing{\spaceskip=0pt\relax}
\providecommand\BIBentryALTinterwordstretchfactor{4}
\providecommand\BIBentryALTinterwordspacing{\spaceskip=\fontdimen2\font plus
\BIBentryALTinterwordstretchfactor\fontdimen3\font minus \fontdimen4\font\relax}
\providecommand\BIBforeignlanguage[2]{{%
\expandafter\ifx\csname l@#1\endcsname\relax
\typeout{** WARNING: IEEEtran.bst: No hyphenation pattern has been}%
\typeout{** loaded for the language `#1'. Using the pattern for}%
\typeout{** the default language instead.}%
\else
\language=\csname l@#1\endcsname
\fi
#2}}

\bibitem{7742315}
O.~Khatib, X.~Yeh, G.~Brantner, B.~Soe, B.~Kim, S.~Ganguly, H.~Stuart, S.~Wang, M.~Cutkosky, A.~Edsinger, P.~Mullins, M.~Barham, C.~R. Voolstra, K.~N. Salama, M.~L'Hour, and V.~Creuze, ``Ocean one: A robotic avatar for oceanic discovery,'' \emph{IEEE Robotics and Automation Magazine}, vol.~23, no.~4, pp. 20--29, 2016.

\bibitem{stuart2017ocean}
H.~Stuart, S.~Wang, O.~Khatib, and M.~R. Cutkosky, ``The ocean one hands: An adaptive design for robust marine manipulation,'' \emph{The International Journal of Robotics Research}, vol.~36, no.~2, pp. 150--166, 2017.

\bibitem{brantner2021controlling}
G.~Brantner and O.~Khatib, ``Controlling ocean one: Human--robot collaboration for deep-sea manipulation,'' \emph{Journal of Field Robotics}, vol.~38, no.~1, pp. 28--51, 2021.

\bibitem{huang2020review}
H.~Huang, Q.~Tang, J.~Li, W.~Zhang, X.~Bao, H.~Zhu, and G.~Wang, ``A review on underwater autonomous environmental perception and target grasp, the challenge of robotic organism capture,'' \emph{Ocean Engineering}, vol. 195, p. 106644, 2020.

\bibitem{aldhaheri2022underwater}
S.~Aldhaheri, G.~De~Masi, {\`E}.~Pairet, and P.~Ard{\'o}n, ``Underwater robot manipulation: Advances, challenges and prospective ventures,'' in \emph{OCEANS 2022-Chennai}.\hskip 1em plus 0.5em minus 0.4em\relax IEEE, 2022, pp. 1--7.

\bibitem{724376}
K.~Leabourne and S.~Rock, ``Model development of an underwater manipulator for coordinated arm-vehicle control,'' in \emph{IEEE Oceanic Engineering Society. OCEANS'98. Conference Proceedings (Cat. No.98CH36259)}, vol.~2, 1998, pp. 941--946 vol.2.

\bibitem{5978927}
S.~T. Wilson, A.~P. Sudheer, and S.~Mohan, ``Dynamic modelling, simulation and spatial control of an underwater robot equipped with a planar manipulator,'' in \emph{2011 International Conference on Process Automation, Control and Computing}, 2011, pp. 1--6.

\bibitem{gumucsel2011modelling}
L.~G{\"u}m{\"u}{\c{s}}el and N.~G. {\"O}zmen, ``Modelling and control of manipulators with flexible links working on land and underwater environments,'' \emph{Robotica}, vol.~29, no.~3, pp. 461--470, 2011.

\bibitem{7002989}
C.~Barbălată, M.~W. Dunnigan, and Y.~Pétillot, ``Dynamic coupling and control issues for a lightweight underwater vehicle manipulator system,'' in \emph{2014 Oceans - St. John's}, 2014, pp. 1--6.

\bibitem{6907522}
E.~Kelasidi, K.~Y. Pettersen, J.~T. Gravdahl, and P.~Liljebäck, ``Modeling of underwater snake robots,'' in \emph{2014 IEEE International Conference on Robotics and Automation (ICRA)}, 2014, pp. 4540--4547.

\bibitem{SVERDRUPTHYGESON201681}
\BIBentryALTinterwordspacing
J.~Sverdrup-Thygeson, E.~Kelasidi, K.~Pettersen, and J.~Gravdahl, ``Modeling of underwater swimming manipulators**this research was partly funded by the research council of norway through the centres of excellence funding scheme, project no. 223254 ntnu amos, and partly funded by vista, a basic research program in collaboration between the norwegian academy of science and letters, and statoil.'' \emph{IFAC-PapersOnLine}, vol.~49, no.~23, pp. 81--88, 2016, 10th IFAC Conference on Control Applications in Marine SystemsCAMS 2016. [Online]. Available: \url{https://www.sciencedirect.com/science/article/pii/S2405896316319127}
\BIBentrySTDinterwordspacing

\bibitem{8121980}
J.~Sverdrup-Thygeson, E.~Kelasidi, K.~Y. Pettersen, and J.~T. Gravdahl, ``The underwater swimming manipulator—a bioinspired solution for subsea operations,'' \emph{IEEE Journal of Oceanic Engineering}, vol.~43, no.~2, pp. 402--417, 2018.

\bibitem{fairchild2023physics}
P.~Fairchild, Y.~Mei, and X.~Tan, ``Physics-informed online estimation of stiffness and shape of soft robotic manipulators,'' \emph{IEEE Control Systems Letters}, 2023.

\bibitem{BARBALATA2018150}
\BIBentryALTinterwordspacing
C.~Barbalata, M.~W. Dunnigan, and Y.~Petillot, ``Position/force operational space control for underwater manipulation,'' \emph{Robotics and Autonomous Systems}, vol. 100, pp. 150--159, 2018. [Online]. Available: \url{https://www.sciencedirect.com/science/article/pii/S092188901730386X}
\BIBentrySTDinterwordspacing

\bibitem{BARBALATA201544}
\BIBentryALTinterwordspacing
C.~Barbălată, M.~W. Dunnigan, and Y.~Pétillot, ``Reduction of the dynamic coupling in an underwater vehicle-manipulator system using an inverse dynamic model approach,'' \emph{IFAC-PapersOnLine}, vol.~48, no.~2, pp. 44--49, 2015, 4th IFAC Workshop onNavigation, Guidance and Controlof Underwater VehiclesNGCUV 2015. [Online]. Available: \url{https://www.sciencedirect.com/science/article/pii/S2405896315002475}
\BIBentrySTDinterwordspacing

\bibitem{HAUGALOKKEN20181}
\BIBentryALTinterwordspacing
B.~O.~A. Haugaløkken, E.~K. Jørgensen, and I.~Schjølberg, ``Experimental validation of end-effector stabilization for underwater vehicle-manipulator systems in subsea operations,'' \emph{Robotics and Autonomous Systems}, vol. 109, pp. 1--12, 2018. [Online]. Available: \url{https://www.sciencedirect.com/science/article/pii/S0921889018300952}
\BIBentrySTDinterwordspacing

\bibitem{mnih2015human}
V.~Mnih, K.~Kavukcuoglu, D.~Silver, A.~A. Rusu, J.~Veness, M.~G. Bellemare, A.~Graves, M.~Riedmiller, A.~K. Fidjeland, G.~Ostrovski, \emph{et~al.}, ``Human-level control through deep reinforcement learning,'' \emph{nature}, vol. 518, no. 7540, pp. 529--533, 2015.

\bibitem{tassa2018deepmind}
Y.~Tassa, Y.~Doron, A.~Muldal, T.~Erez, Y.~Li, D.~d.~L. Casas, D.~Budden, A.~Abdolmaleki, J.~Merel, A.~Lefrancq, \emph{et~al.}, ``Deepmind control suite,'' \emph{arXiv preprint arXiv:1801.00690}, 2018.

\bibitem{jadhav2017rov}
Y.~R. Jadhav and Y.~S. Moon, ``Rov manipulation from observation and exploration using deep reinforcement learning,'' \emph{Journal of Advanced Research in Ocean Engineering}, vol.~3, no.~3, pp. 136--148, 2017.

\bibitem{Carlucho_2020}
\BIBentryALTinterwordspacing
I.~Carlucho, M.~De~Paula, C.~Barbalata, and G.~G. Acosta, ``A reinforcement learning control approach for underwater manipulation under position and torque constraints,'' in \emph{Global Oceans 2020: Singapore – U.S. Gulf Coast}.\hskip 1em plus 0.5em minus 0.4em\relax IEEE, Oct. 2020. [Online]. Available: \url{http://dx.doi.org/10.1109/IEEECONF38699.2020.9389378}
\BIBentrySTDinterwordspacing

\bibitem{yang2021prediction}
H.~Yang, J.~Liu, X.~Fang, X.~Chen, Z.~Gong, S.~Wang, S.~Kong, J.~Yu, and L.~Wen, ``Prediction model-based learning adaptive control for underwater grasping of a soft manipulator,'' \emph{International Journal of Intelligent Robotics and Applications}, vol.~5, pp. 337--353, 2021.

\bibitem{8729801}
A.~Nikou, C.~K. Verginis, and D.~V. Dimarogonas, ``A tube-based mpc scheme for interaction control of underwater vehicle manipulator systems,'' in \emph{2018 IEEE/OES Autonomous Underwater Vehicle Workshop (AUV)}, 2018, pp. 1--6.

\bibitem{salloom2020adaptive}
T.~Salloom, X.~Yu, W.~He, and O.~Kaynak, ``Adaptive neural network control of underwater robotic manipulators tuned by a genetic algorithm,'' \emph{Journal of Intelligent \& Robotic Systems}, vol.~97, pp. 657--672, 2020.

\bibitem{carlucho2021adaptive}
I.~Carlucho, D.~W. Stephens, and C.~Barbalata, ``An adaptive data-driven controller for underwater manipulators with variable payload,'' \emph{Applied Ocean Research}, vol. 113, p. 102726, 2021.

\bibitem{oubre2021datadrivencontrollersneedperception}
\BIBentryALTinterwordspacing
J.~P. Oubre, I.~Carlucho, and C.~Barbalata, ``Data-driven controllers and the need for perception systems in underwater manipulation,'' 2021. [Online]. Available: \url{https://arxiv.org/abs/2109.10327}
\BIBentrySTDinterwordspacing

\bibitem{chee2023learnest}
K.~Y. Chee and M.~A. Hsieh, ``{LEARNEST}: {LEARN}ing {E}nhanced {M}odel-based {S}tate {EST}imation for robots using knowledge-based neural ordinary differential equations,'' in \emph{2023 IEEE International Conference on Robotics and Automation (ICRA)}.\hskip 1em plus 0.5em minus 0.4em\relax IEEE, 2023, pp. 11\,590--11\,596.

\bibitem{rahman2022svin2}
S.~Rahman, A.~Quattrini~Li, and I.~Rekleitis, ``Svin2: A multi-sensor fusion-based underwater slam system,'' \emph{The International Journal of Robotics Research}, vol.~41, no. 11-12, pp. 1022--1042, 2022.

\bibitem{davison2007monoslam}
A.~J. Davison, I.~D. Reid, N.~D. Molton, and O.~Stasse, ``Monoslam: Real-time single camera slam,'' \emph{IEEE transactions on pattern analysis and machine intelligence}, vol.~29, no.~6, pp. 1052--1067, 2007.

\bibitem{matsuki2024gaussian}
H.~Matsuki, R.~Murai, P.~H. Kelly, and A.~J. Davison, ``Gaussian splatting slam,'' in \emph{Proceedings of the IEEE/CVF Conference on Computer Vision and Pattern Recognition}, 2024, pp. 18\,039--18\,048.

\bibitem{kerbl20233d}
B.~Kerbl, G.~Kopanas, T.~Leimk{\"u}hler, and G.~Drettakis, ``3d gaussian splatting for real-time radiance field rendering.'' \emph{ACM Trans. Graph.}, vol.~42, no.~4, pp. 139--1, 2023.

\bibitem{mildenhall2021nerf}
B.~Mildenhall, P.~P. Srinivasan, M.~Tancik, J.~T. Barron, R.~Ramamoorthi, and R.~Ng, ``Nerf: Representing scenes as neural radiance fields for view synthesis,'' \emph{Communications of the ACM}, vol.~65, no.~1, pp. 99--106, 2021.

\bibitem{liu2023zero}
R.~Liu, R.~Wu, B.~Van~Hoorick, P.~Tokmakov, S.~Zakharov, and C.~Vondrick, ``Zero-1-to-3: Zero-shot one image to 3d object,'' in \emph{Proceedings of the IEEE/CVF international conference on computer vision}, 2023, pp. 9298--9309.

\bibitem{wang2024dust3r}
S.~Wang, V.~Leroy, Y.~Cabon, B.~Chidlovskii, and J.~Revaud, ``Dust3r: Geometric 3d vision made easy,'' in \emph{Proceedings of the IEEE/CVF Conference on Computer Vision and Pattern Recognition}, 2024, pp. 20\,697--20\,709.

\bibitem{tang2024uwnerf}
Y.~Tang, C.~Zhu, R.~Wan, C.~Xu, and B.~Shi, ``Neural underwater scene representation,'' 2024.

\bibitem{zhang2023beyond}
T.~Zhang and M.~Johnson-Roberson, ``Beyond nerf underwater: Learning neural reflectance fields for true color correction of marine imagery,'' \emph{IEEE Robotics and Automation Letters}, 2023.

\bibitem{6301026}
H.~Durrant-Whyte, N.~Roy, and P.~Abbeel, \emph{Learning to Control a Low-Cost Manipulator Using Data-Efficient Reinforcement Learning}, 2012, pp. 57--64.

\bibitem{Kober2014}
J.~Kober and J.~Peters, \emph{Movement Templates for Learning of Hitting and Batting}.\hskip 1em plus 0.5em minus 0.4em\relax Cham: Springer International Publishing, 2014, pp. 69--82.

\bibitem{6095096}
M.~Kalakrishnan, L.~Righetti, P.~Pastor, and S.~Schaal, ``Learning force control policies for compliant manipulation,'' in \emph{2011 IEEE/RSJ International Conference on Intelligent Robots and Systems}, 2011, pp. 4639--4644.

\bibitem{levine2016end}
S.~Levine, C.~Finn, T.~Darrell, and P.~Abbeel, ``End-to-end training of deep visuomotor policies,'' \emph{Journal of Machine Learning Research}, vol.~17, no.~39, pp. 1--40, 2016.

\bibitem{schulman2017proximal}
J.~Schulman, F.~Wolski, P.~Dhariwal, A.~Radford, and O.~Klimov, ``Proximal policy optimization algorithms,'' \emph{arXiv preprint arXiv:1707.06347}, 2017.

\bibitem{haarnoja2018soft}
T.~Haarnoja, A.~Zhou, K.~Hartikainen, G.~Tucker, S.~Ha, J.~Tan, V.~Kumar, H.~Zhu, A.~Gupta, P.~Abbeel, \emph{et~al.}, ``Soft actor-critic algorithms and applications,'' \emph{arXiv preprint arXiv:1812.05905}, 2018.

\bibitem{akkaya2019solving}
I.~Akkaya, M.~Andrychowicz, M.~Chociej, M.~Litwin, B.~McGrew, A.~Petron, A.~Paino, M.~Plappert, G.~Powell, R.~Ribas, \emph{et~al.}, ``Solving rubik's cube with a robot hand,'' \emph{arXiv preprint arXiv:1910.07113}, 2019.

\bibitem{ha2018world}
D.~Ha and J.~Schmidhuber, ``World models,'' \emph{arXiv preprint arXiv:1803.10122}, 2018.

\bibitem{hafner2019learning}
D.~Hafner, T.~Lillicrap, I.~Fischer, R.~Villegas, D.~Ha, H.~Lee, and J.~Davidson, ``Learning latent dynamics for planning from pixels,'' in \emph{International conference on machine learning}.\hskip 1em plus 0.5em minus 0.4em\relax PMLR, 2019, pp. 2555--2565.

\bibitem{hafner2019dream}
D.~Hafner, T.~Lillicrap, J.~Ba, and M.~Norouzi, ``Dream to control: Learning behaviors by latent imagination,'' \emph{arXiv preprint arXiv:1912.01603}, 2019.

\bibitem{wu2023daydreamer}
P.~Wu, A.~Escontrela, D.~Hafner, P.~Abbeel, and K.~Goldberg, ``Daydreamer: World models for physical robot learning,'' in \emph{Conference on robot learning}.\hskip 1em plus 0.5em minus 0.4em\relax PMLR, 2023, pp. 2226--2240.

\bibitem{liang2024dreamitate}
J.~Liang, R.~Liu, E.~Ozguroglu, S.~Sudhakar, A.~Dave, P.~Tokmakov, S.~Song, and C.~Vondrick, ``Dreamitate: Real-world visuomotor policy learning via video generation,'' \emph{arXiv preprint arXiv:2406.16862}, 2024.

\bibitem{zeng2020tossingbot}
A.~Zeng, S.~Song, J.~Lee, A.~Rodriguez, and T.~Funkhouser, ``Tossingbot: Learning to throw arbitrary objects with residual physics,'' \emph{IEEE Transactions on Robotics}, vol.~36, no.~4, pp. 1307--1319, 2020.

\bibitem{xu2022dextairity}
Z.~Xu, C.~Chi, B.~Burchfiel, E.~Cousineau, S.~Feng, and S.~Song, ``Dextairity: Deformable manipulation can be a breeze,'' \emph{arXiv preprint arXiv:2203.01197}, 2022.

\bibitem{ha2022flingbot}
H.~Ha and S.~Song, ``Flingbot: The unreasonable effectiveness of dynamic manipulation for cloth unfolding,'' in \emph{Conference on Robot Learning}.\hskip 1em plus 0.5em minus 0.4em\relax PMLR, 2022, pp. 24--33.

\bibitem{chi2024iterative}
C.~Chi, B.~Burchfiel, E.~Cousineau, S.~Feng, and S.~Song, ``Iterative residual policy: for goal-conditioned dynamic manipulation of deformable objects,'' \emph{The International Journal of Robotics Research}, vol.~43, no.~4, pp. 389--404, 2024.

\bibitem{liu2024paperbot}
R.~Liu, J.~Liang, S.~Sudhakar, H.~Ha, C.~Chi, S.~Song, and C.~Vondrick, ``Paperbot: Learning to design real-world tools using paper,'' \emph{arXiv preprint arXiv:2403.09566}, 2024.

\bibitem{chi2023diffusion}
C.~Chi, S.~Feng, Y.~Du, Z.~Xu, E.~Cousineau, B.~Burchfiel, and S.~Song, ``Diffusion policy: Visuomotor policy learning via action diffusion,'' \emph{arXiv preprint arXiv:2303.04137}, 2023.

\bibitem{zhao2023learningfinegrainedbimanualmanipulation}
\BIBentryALTinterwordspacing
T.~Z. Zhao, V.~Kumar, S.~Levine, and C.~Finn, ``Learning fine-grained bimanual manipulation with low-cost hardware,'' 2023. [Online]. Available: \url{https://arxiv.org/abs/2304.13705}
\BIBentrySTDinterwordspacing

\bibitem{lee2024behavior}
S.~Lee, Y.~Wang, H.~Etukuru, H.~J. Kim, N.~M.~M. Shafiullah, and L.~Pinto, ``Behavior generation with latent actions,'' \emph{arXiv preprint arXiv:2403.03181}, 2024.

\end{thebibliography}

\end{document}